# Vietnamese Open Information Extraction

Diem Truong
Faculty of Information Technology
Angiang University, Vietnam
ttdiem@agu.edu.vn

Duc-Thuan Vo*
Ryerson University
Toronto, ON, Canada
thuanvd@gmail.com

Uyen Trang Nguyen
York University
Toronto, ON, Canada
utn@cse.yorku.ca

## ABSTRACT

Open information extraction (OIE) is the process to extract relations and their arguments automatically from textual documents without the need to restrict the search to predefined relations. In recent years, several OIE systems for the English language have been created but there is not any system for the Vietnamese language. In this paper, we propose a method of OIE for Vietnamese using a clause-based approach. Accordingly, we exploit Vietnamese dependency parsing using grammar clauses that strives to consider all possible relations in a sentence. The corresponding clause types are identified by their propositions as extractable relations based on their grammatical functions of constituents. As a result, our system is the first OIE system named vnOIE [1] for the Vietnamese language that can generate open relations and their arguments from Vietnamese text with highly scalable extraction while being domain independent. Experimental results show that our OIE system achieves promising results with a precision of 83.71%.

## CCS CONCEPTS

I.2.7 [**Computing Methodologies**]: Artificial intelligence → Natural language processing

## 1 INTRODUCTION

Relation Extraction (RE) is one of the important tasks of natural language processing (NLP).The goal of relation extraction is to discover the relevant segments of textual documents so that they can present unstructured data in the form of structured relations. Open Information Extraction (OIE) is a more general approach of RE which relies minimally on background knowledge and manually labeled training data [5-6, 28, 30]. In OIE, various types of relations are taken into consideration without the need to restrict the search to pre-specified relations. Some recent OIE techniques use handcrafted extraction heuristics or automatically constructed training data to learn extractors or to estimate the confidence of propositions. Banko et al. [1], Wu et al. [31] and Fader et al. [7] obtained a shallow syntactic representation of natural language text in the form of verbs or verbal phrases and their arguments. Other approaches [4, 11, 29, 31] used dependency parsing for OIE. These approaches use various heuristics to obtain relation propositions based on dependency analysis.

In Vietnamese NLP, research has been conducted on fundamental tasks such as Part-of-Speech (POS) [15], Chunking [13], and Parsing [16, 25] while simultaneously other approaches are dealing on specified tasks such as Named Entity Recognition (NER), Semantic Role Labeling (SRL) [18] and RE [20, 21]. With respect to RE, most proposed techniques in particular recognized the relations between entities in order to satisfy predefined target relations. Sam et al. [20] used conditional random field with various features of entities such as entity types and entity index to train relations in the sentence. Tran et al. [26] used seed relations to learn patterns on the repeated processes of refined patterns where the relations could be identified with the best match patterns in a question answering system. Sam et al. [21] exploited bag-of-word, POS and entity types to extract relations using a bootstrapping method. The authors used shallow linguistic kernel functions to combine relations and two-candidate entities, and then trained them using a bootstrapping approach. These approaches not only face the problem of requiring a large hand-annotation corpus, which takes large amounts of example data and expertise to label data or training data, but also limit the targeted domain. In this paper, we propose a system named vnOIE for Vietnamese open information extraction. Essentially, the system extracts relations and their arguments from Vietnamese text based on dependency parsing and grammar clauses. We focus on identifying the set of clauses in each sentence. For each clause, the corresponding clause type is determined in accordance to the grammatical function of its coherent constituent for extracting relation tuples. To the best of our knowledge, our system is the first OIE system for Vietnamese.

---
* Corresponding author.
[1] https://bitbucket.org/vnOIE

The remainder of this paper is organized as follows. Section 2 offers overview of relevant research and explores methods of OIE that provide further insight into the issues addressed in the current study. In Section 3, we present a detailed description of Vietnamese dependency parsing and grammar clauses that lead to the proposed clause-based method for the vnOIE system. Section 4 presents experimental results and discussion of our vnOIE system. Finally, we conclude the paper and outline future work in the last section.

## 2 RELATED WORK

Banko et al. [1] presented TextRunner considered to be the pioneering OIE system for English. In recent years, most of existing OIE work exploited the presentation of verbs and their arguments to improve on TextRunner based on shallow syntactic [7, 31] and dependency parsing [4, 7, 11, 29]. In this section, we discuss on approaches that are related to OIE.

Several OIE systems exploited syntactic parsing for relation extraction. TextRunner [1] and ReVerb [7] used training data and syntactic analysis while WOE$^{pos}$ trained the corpus automatically by obtaining infoboxes from Wikipedia. TextRunner trained a Bayes classifier in an offline phase and applies it for the efficient extraction of propositions in the online phase. Fader et al. [7] have developed ReVerb, which made use of syntactical and lexical constraints to solve the problem of incoherent and uninformative extractions. Instead of extracting entities first, ReVerb extracts verbal relation sequences based on a set of POS patterns. Then entities are identified around the relation sequences, so the system only extracts relation tokens between two entities. WOE$^{pos}$ [31] also used a classifier, but the classifier is based on a high-quality training corpus obtained automatically from the infoboxes of Wikipedia for improved precision and recall.

On the other hand, several systems focus on the use of dependency parsing for OIE. WOE$^{parse}$ [31] uses automatically generated training data to learn extraction patterns on dependency parsing. Mausam et al. [11] presented OLLIE, which uses hand-labeled data to create a training set which includes millions of relations extracted by ReVerb. OLLIE learns relation patterns from the dependency path and lexicon information such that the relations that match the identified patterns will be extracted. A more recent OIE system, ClausIE [4], LS3RyIE [29], uses dependency parsing and a small set of domain-independent lexica without any post-processing or training data. It exploits linguistic knowledge about the grammar of the English language to first detect clauses in an input sentence and to subsequently identify the type of each clause according to the grammatical function of its constituents. Therefore, these systems are able to generate high-precision extractions and can be flexibly customized to the underlying application domain.

Other OIE systems [8, 12, 22] apply amount of knowledge as Wikipedia, DBpedia or Freebase as distant supervision to solve NER ambiguity for OIE. Mirezaei et al. [12] proposed to complement systems like ReVerb or OLLIE, which often fails to extract nouns and mediated relations, with an approach based on collecting patterns from text by using Wikipedia, DBpedia, and Freebase to build sets of sentences illustrating the occurrences of particular relations. Garcia at al. [8] leveraged dependency parsing in order to simplify the linguistic structure of a sentence, and then applied semantic extraction rules, obtained via distant supervision, over these results. Schmidek et al. [22] proposed to simplify sentences by leveraging clauses which are already tagged NER by dependency parsing. The authors used a supervised method, Naive Bayes classifier, to detect dependency component chunking in order to connect or disconnect them as relation association in a sentence.

As mentioned earlier, RE in the Vietnamese faces the problem of requiring a large hand annotated corpus and limits the fixed relations in specified domains. In this paper, we applied the work presented by Corro et al. [4] for generating relations from grammar clauses by exploiting DP structures. In particular, we identify the set of clauses in each sentence. For each clause, we identify the corresponding clause type according to the grammatical function of its coherent constituent as extractable relations. To the best of our knowledge, our system is the first system for Vietnamese OIE.

**Table 1: CoNLL format of the sentence "Tiến_sĩ Minh là hiệu_trưởng TrườngĐại_họcAn_Giang /Dr. Minh is the president of Angiang University"**

| ID | FORM | CPOSTAG | POSTAG | HEAD | DEPREL |
|---|---|---|---|---|---|
| 1 | Tiến_sĩ/Dr. | N | Nc | 3 | Sub |
| 2 | Minh | N | Np | 1 | nmod |
| 3 | là/is | V | V | 0 | root |
| 4 | hiệu_trưởng/the president of | N | N | 3 | dob |
| 5 | Trường | N | N | 4 | loc |
| 6 | Đại học/ university | N | N | 5 | nmod |
| 7 | An_giang/ Angiang | N | Np | 5 | nmod |
| 8 | . | . | . | 3 | punct |

## 3 PROPOSED METHOD

Basically, an OIE system extracts relation tuples representing basic clauses or assertions from text. In this context, clauses are defined as coherent and non-over-specified pieces of basic information [19]. Like English, a Vietnamese clause [10] is unit of grammatical organization consisting of a subject and predicate. It is considered as part of a sentence that expresses some coherent component of information [3]. Following [4], we exploit grammatical clauses and dependency parsing in order to extract relation triples of interest in Vietnamese documents. We begin on this process by recognizing a set of clauses to derive a set of coherent constituents for generating their propositions. When the clauses are determined that can produce correct extractable relations. Vietnamese dependency parsing is used to identify clauses in our method.

### 3.1 Vietnamese Dependency Parsing

Several methods have developed for Vietnamese DP in recent years [9, 16, 25]. Hong et al. [9] presented a method to extract dependency relations based on lexicalized tree-adjoining grammar for DP. Thi et al. [25] used head-percolation rules to find the head of each constituent phrase to transform from constituency structure into dependency structure. Nguyen et al. [16] automatically created DP from a constituent structure-based Vietnamese Treebank. Most of these methods presented DP based on standard CoNLL structure [2, 17]. With such an organization, tokens are combined with each other for their dependence relationship. In Vietnamese, DP is commonly presented in six fields such as ID, FORM, CPOSTAG, POSTAG, HEAD and DEPREL shown in Table 1. Given the sentence "Tiến_sĩ Minh là hiệu_trưởng Trường Đại_học An_Giang/Dr. Minh is the president of Angiang University", the subject "Tiến_sĩ Minh/Dr. Minh" is defined as main subject which is associated with predicate "là/is" via *sub*(Tiến_sĩ/Dr., là/is ) and *nmod* (Tiến_sĩ/Dr., Minh).

Table 2: Vietnamese clause types [3, 10]

| Clause types | Patterns | Sentences | Derived clauses |
|---|---|---|---|
| SV | SV<br>SVA<br>SVA<br>SVAA | Minh dạy_học ở Trường Đại_học An_Giang từ năm 2010. / Minh has taught at Angiang University since 2010. | SV: ("Minh/Minh", "dạy_học/has taught")<br>SVA: ("Minh", " dạy_học/has taught", "ở Trường Đại_học An_Giang/at Angiang university")<br>SVA: ("Minh", " dạy_học/has taught ", "từ năm 2010/since 2010")<br>SVAA: ("Minh", " dạy_học /has taught", "ở Trường Đại_học An_Giang/at Angiang university", "từ năm 2010/since 2010") |
| SVA | SVA<br>SVAA | Minh đến thư viện hôm qua./Minh came to the library yesterday. | SVA: ("Minh", "đến/came to", "thư viện/the library")<br>SVAA: ("Minh", "đến/came to", "thư viện/the library", "hôm qua/yesterday") |
| SVC | SVC<br>SVCA | Minh là giảng viên ở Trường Đại_học An_Giang/Minh is a lecturer at Angiang University. | SVC: ("Minh", "là/is", "giảng_viên/lecturer")<br>SVCA: ("Minh", "là/is", "giảng_viên/lecturer", "ở trường Đại_học An_Giang/atAngiang University") |
| SVO | SVO<br>SVOA | Minh đang đọc sách trong thư viện./Minh is reading a book at the library. | SVO: ("Minh", "đang đọc/is reading", "sách/a book")<br>SVOA: ("Minh", "đang đọc/is reading", "sách/abook", "trong thư viện/at the library") |
| SVOO | SVOO | Minh dạy tiếng Anh cho sinh_viên./ Minh teaches English for students. | SVOO: ("Minh", "dạy/teaches", "tiếng Anh/English", "cho sinh_viên/for students") |
| SVOA | SVOA | Minh để quyển sách trên bàn/Minh puts the book on the table. | SVOA: ("Minh", "để/puts", "quyển sách/a book", "trên bàn/on the table") |
| SVOC | SVOC<br>SVOCA | Minh tìm thấy quyển sách thú_vị trong thư viện/Minh found an interesting book in the library. | SVOC: ("Minh", "tìm thấy/found", "quyển sách/a book", "thú_vị/interesting")<br>SVOCA: ("Minh", "tìm thấy/found", "quyển sách/a book", "thú_vị/interesting","trong thư viện/in the library") |

### 3.2 Clause-based Vietnamese OIE

In Vietnamese, a sentence is a grammatical unit consisting of one or more clauses. A clause consists of different components including subject (S), verb (V), indirect object (O), direct object (O), complement (C), and/or one or more adverbs (A) that present a subject and predicate [3]. According to [10], Vietnamese grammar basically has seven clause types corresponding to seven clause types in English, which are presented in the forms SV, SVA, SVC, SVO, SVOO, SVOA, and SVOC shown in Table 2.

For instance, the clause type for the sentence "Minh đến thư_viện hôm qua./Minh came to the library yesterday" is SVA with subject, "Minh", verb, "đến/came" and adverb, "thưviện/library". vnOIE system is based on seven clause types and their coherence that plays to OIE. The system recognizes sets of clause types from the input sentence based on the DP analysis as introduced earlier. Then, number of clause types will be determined based on the quality of verbs in the sentence [23]. For Example, a sentence "Tiến_sĩ Minh là một giáo_viên và thường giúp_đỡ nhiều sinh_viên nghèo ở Trường Đại_học An_Giang/Dr. Minh is a lecture and often helps many poor students in Angiang

University" is presented by SVC ("Tiến_sĩ Minh/Dr.Minh", "là/is", "giảng_viên/a lecturer") with verb, "la`/is" and SVO("Tiến_sĩ Minh/Dr. Minh", "thường giúp_đỡ/often helps", "nhiều sinh_viên nghèo/many poor students") with verb, "thường giúp_đỡ/often helps". When a clause is identified, a set of coherently derived-clauses will be determined based on the constituents of the clause examples shown in Table 2. vnOIE exploits clauses for the purpose of extracting relation extractions using the following two steps (1) Detecting clause types and (2) Extracting relations.

---

**Algorithm 1**: Identifying clause types in a sentence [4, 29]
**Input:** DP from a sentence
**Output:** Set of clause types

1:  S, V, C, O, A = null
         // S: Subject, V: Verb,
         // C: Complement, O: Object, A: Adverb
2:  **if** found(S, V) **do**
3:       C ← Constituents (C)
4:       O ← Constituents (O)
5:       A ← Constituents (A)
6:  **end if**
7:  **if** direct(O) **do**
8:       **if** indirect(O) **do**
9:            ClauseTypes ← SVOO
10:      **else if** found(A) **do**
11:           ClauseTypes ← SVOA
12:      **else if** found(C) **do**
13:           ClauseTypes ← SVOC
14:           **else** ClauseTypes ← SVO
15: **else if** found(C) **do**
16:          ClauseTypes ← SVC
17:     **else if** found(A) **do**
18:          ClauseTypes ← SVA
19:          **else** ClauseTypes ← SV
20: **return** ClauseTypes

---

**Detecting Clause Types.** According to Tinh [23] the number of verbs in a sentence indicates the number of clause types in the sentence. Following [4, 29], vnOIE detects clauses in a sentence based on the grammatical clause structures shown in Algorithm 1. The algorithm will find the subjects and the organization of the verb via *nsubj*, *compound*, *coord*, *conj* and *vmod* from DP. When the subject and verb(s) are found, the algorithm will seek all associated constituents with main clause types following found verb(s) such as constituents of object (O), constituents of complement (C), and constituents of adverb (A) presented in lines 3-5. Then the algorithm will indentify clause types as SVO, SVC or SVA in lines 14, 16, and 18, respectively. Moreover, the algorithm will seek direct/indirect objects for generating the SVOO clause type as in Line 9 or will seek complement C for generating the clause SVOC in line 13. Otherwise, the algorithm will seek adverb A for generating SVOA as in line 11 of the algorithm. Moreover, clause types SVC, SVOO, and SVOC are identified solely based on the structure of the clause. All adverbs are optional in all clauses. For the examples shown in Table 2, the derived clauses from SVO are SVO and SVOA, or the derived clauses from SVA are SVA and SVAA. Optional adverbs "A!" and "A?" indicate essential adverbs and optional adverbs, respectively. Clause types SVC, SVOO, and SVOC are identified solely by the structure of the clause; all adverbs are optional for these types as SVCA, SVOA and SVOCA.

**Extracting relations.** When clause types are indentified, vnOIE will extract relations based on patterns of clause types shown in Table 1. Each pattern consists of all the constituents of the clause that present a subject, a relation and one or more arguments. To generate a proposition as a relation triple (*arg1*, *rel*, *arg2*), the subject is considered as the first argument (*arg1*) of each clause; the verb, as relation (*rel*). They are then used to construct the proposition. For example, for the clause type SV in Table1, the subject "Minh" and the verb "dạy/has taught" of the clause are used to construct the proposition with the following list of patterns: SV, SVA, and SVAA. The remaining argument (*arg2*) such as object, complement and adverb will be determined by the connection with the relation. Consequently, the system combines all arguments in the propositions in order to extract triple relations. For example, for the sentence "Minh dạy_học ở Trường Đại_học An_Giang từ năm 2010./Minh has taught at Angiang University since 2010.", four coherent relations are derived as follows:

- SV: (S: "Minh", V: "dạy_học/has taught")
- SVA: (S: "Minh", V: "dạy_học ở/has taught at", A: "Trường Đại_học An_Giang/Angiang university")
- SVA: (S: "Minh", V: "dạy_học từ/has taught since", "2010")
- SVAA: (S: "Minh", V: "dạy_học /has taught ở", A: "trường Đại_học An_Giang/at Angiang university", A: "từ năm 2010/since 2010")

## 4  EXPERIMENTATION

In this section, we first describe the settings of our experiments, followed by experimental results. We then discuss analyze on sample extractions and errors in vnOIE system.

### 4.1  Experimental setting

In this study, we conducted experiments on the Vietnamese dependency TreeBank corpus [16], which has been annotated with CONLL format as DP input. The corpus contains 10,197 sentences with 218,749 tokens. Similarly to other OIE [4], our system does not process question sentences. After removing these sentences, 7,767 remaining sentences consisting of 152,396 tokens are used for experiments. The outputs of our system are labeled and verified manually by two independent experts. To guarantee the accuracy of system's performance, each extraction was labeled based on the description of the original TreeBank corpus. The experts were instructed to treat an extraction as correct if it was both correct and informative. The extractions that lacked meaning were labeled as incorrect. Particularly, the correct extractions have to be approved and labeled as correct by both experts. Two experts independently evaluated the system outputs. If a clause is deemed correctly extracted by both experts then the result will be tagged as "Correct". Otherwise, it will be tagged as "Incorrect". In cases where one expert agrees with the extraction but the other does not then the result will be tagged as "Incorrect". vnOIE is domain-independent OIE in the sense that the extraction is not fixed with any pre-defined Vietnamese relations. Thus, precision is used for evaluating the performance of the system.

### 4.2 Results

We evaluate vnOIE based on several categories such as the number of verbs and number of extracted clauses in a sentence.

Table 3 shows the statistical output of vnOIE for seven cases based on the number of verbs ranging from 1 to 6 and larger than 6 in a sentence. A sentence with a higher number of verbs will lead to a higher number of generated clauses. For example, 8996 clauses are extracted from 2401 sentences containing two verbs while 1672 clauses are produced from only 136 sentences having six verbs. Figure 1 depicts the results of vnOIE based on seven cases of the number of verbs. We can see that the lower the number of verbs in a sentence, the higher the precision. The system delivers the best result in the case of one-verb clause, with a precision of 92.78%. When clauses have more than six verbs each, the achieved precision is 75.20%. In general, the precision achieved by vnOIE is over 83% when clauses have four verbs or fewer, which is the case for most sentences written/spoken in practice.

**Table 3: Detail of clause generation**

| #verbs | #sentences | #average tokens in a sentence | #clause outputs |
|---|---|---|---|
| 1 | 2612 | 14 | 4361 |
| 2 | 2401 | 19 | 8996 |
| 3 | 1436 | 22 | 8397 |
| 4 | 718 | 26 | 5824 |
| 5 | 357 | 31 | 3611 |
| 6 | 136 | 35 | 1672 |
| >6 | 107 | 42 | 1734 |

With respect to the effectiveness of clause extraction, we evaluate the system based on the number of extracted clauses in a sentence. We evaluate the system by grouping sentences into four categories based on sentence structures, namely, simple sentence, complex sentence, highly complex sentence and extremely complex sentence. A simple sentence produces one clause and a complex sentence, 2-3 clauses. In highly complex sentences, the numbers of generated clauses are 4-6 clauses, while an extremely complex sentence could produce more than six clauses. Table 4 depicts the performance results of the system in each category. The vnOIE system obtains a precision of 92.78%, 84.23%, 80.68, and 75.20% for simple, complex, highly complex and extremely complex sentences, respectively. These numbers indicate that our system succeeds in extracting clauses from most sentences written/spoken in practice, which belong to the first three categories. Finally, vnOIE achieves an overall average precision of 83.71%.

**Table 4: Experimental results on four cases of sentence structure**

| No. | Categories | Ratio (%) | Precision (%) |
|---|---|---|---|
| 1 | Simple | 33.63 | 92.78 |
| 2 | Complex | 49.40 | 84.23 |
| 3 | Highly complex | 15.59 | 80.68 |
| 4 | Extremely complex | 1.38 | 75.20 |
| Overall | | | 83.71 |

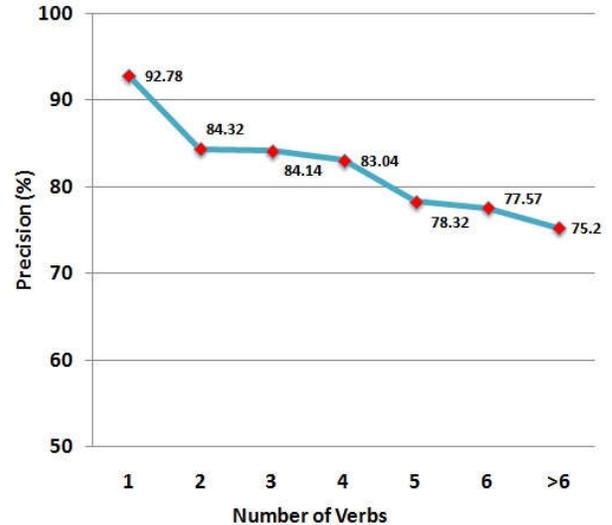

**Figure 1: Experimental results of seven categories.**

### 4.3 Error analysis and sample outputs

vnOIE is the first Vietnamese OIE system which is based on grammatical clauses of the Vietnamese language, highly scalable in terms of clause extraction, and domain independent. The system exploits DP analysis to quickly extract relation tuples based on grammatical clauses. However, some unexpected incorrect extractions could result from the output of Vietnamese sparing. Unlike English DP, DP in Vietnamese may not be able to detect the details of auxiliary adverbs, which could result in incorrect extractions caused by wrongly labeled main verbs. For instance, consider the sentence "Minh được tặng quà trong ngày sinh_nhật/Minh was given a present on birthday" where "được/was" is connected to "tặng/given" via *vmod*(được/was, tặng/given). In this case, the main verb was determined wrongly by *root*(root, được/was) instead of the correct *root*(root, tặng/given). The problem was due to vnOIE using heuristic rules to find significant verbs in a sentence based on DP. Secondly, Vietnamese DP has a limitation on distinguishing between essential adverbs and optional adverbs. Essential adverbs are required, while optional adverbs may or may not appear in extracted clauses. In some cases, vnOIE failed to determine clauses such as SV(A), SVA(A), SVO(A), where components A are essential adverbs.

**Table 5: Sample outputs of six cases of sentence structures with Correct: 1 and Incorrect: 0**

| #verbs | Sentences | Label |
|---|---|---|
| 1 | **Sentence 1:** Hôm_nay, hội_nghị tiếp_tục làm_việc./Today, the conference continues to work. | |
| | p1: ("hội_nghị/the conference", "tiếp_tục/continues làm_việc/to work") | 1 |
| | p2: ("hội_nghị/the conference", "tiếp_tục làm_việc/continues to work", "Hôm_nay/today") | 1 |
| 2 | Sentence 2: Chị chấp_nhận chúng_ta thương_lượng với nhau./She accepted us negotiating together. | |
| | p3: ("Chị/She", "chấp_nhận/accepted") | 1 |
| | p4: ("Chị/She", "chấp_nhận/accepted", "chúng_ta/us thương_lượng/negotiate với nhau/together") | 1 |
| | p5: ("chúng_ta/we", "thương_lượng/negotiate", "với nhau/together") | 1 |
| 3 | **Sentence 3:** Tuấn thường sử_dụng ĐTDĐ, đi xe FX, thích đua xe./Tuan often uses mobile, rides FX, likes racing. | |
| | p6: ("Tuấn", "thường sử_dụng/often uses", "ĐTDĐ/mobile") | 1 |
| | p7: ("Tuấn", "thường sử_dụng/oftern uses", "ĐTDĐ/mobile", "đi xe FX/riding FX") | 1 |
| | p8: ("Tuấn", "đi/rides", "xe FX/FX") | 1 |
| | p9: ("Tuấn", "thường sử_dụng/often uses", "ĐTDĐ/Mobile", "thích đua xe/liking racing ") | 1 |
| | p10: ("Tuấn", "thích/likes", "đua xe/racing") | 1 |
| 4 | **Sentence 4:** Má nghe tin , đau_buồn đến không ăn cơm được nhưng vẫn phải xách giỏ ra chợ , vẫn phải nói cười./Mama hears this news, be sad to not eat meals but she still carries a bag to the market, still be laughing. | |
| | p11: ("Má/Mama", "nghe/hears", "tin/news") | 1 |
| | p12: ("Má/Mama", "nghe/hears", "tin/news", "đau_buồn đến không ăn cơm được/be sad to not eat meals") | 1 |
| | p13: ("Má/Mama", "đau_buồn đến/be sad to") | 0 |
| | p14: ("Má/Mama", "nghe/hears", "tin/news", "nhưng vẫn phải xách giỏ ra chợ/but carrying a bag to the market") | 1 |
| | p15: ("Má/Mama", "vẫn phải xách giỏ/still carry a bag ", "chợ/market") | 0 |
| | p16: ("Má/Mama", "vẫn phải xách giỏ/still carry a bag", "chợ/market", "ra chợ/to market") | 0 |
| | p17: ("Má/Mama", "nghe/hears", "tin/news", "vẫn phải nói cười/still having laugh") | 1 |
| | p18: ("Má/Mama", "vẫn phải nói cười/still have to laugh") | 1 |
| 5 | **Sentence 5:** Ông quả_quyết rằng nếu đây là sự_thật thì ông sẽ ra làng , xã làm đơn từ con và coi như anh Đại không có trên cuộc_đời này./He asserts that if this is a truth, he will go to the village to make a petition for disavowing his son and consider that Mr Dai does not be existed in his life. | |
| | p19: ("Ông/He", "quả_quyết/asserts") | 0 |
| | p20: ("Ông/He", "quả_quyết/asserts", "rằng nếu đây là sự_thật thì ông sẽ ra làng , xã làm đơn từ con và coi như anh Đại không có trên cuộc_đời này/ if this is a truth, he will go to the village to make a petition for disavowing his son and consider that Mr Dai does not be existed in his life") | 1 |
| | p21: ("đây/this", "là/is", "sự_thật/the truth") | 1 |
| | p22: ("đây/this", "là/is", "sự_thật/the truth", "thì ông sẽ ra làng , xã làm đơn từ con và coi như anh Đại không có trên cuộc_đời này/that he will go to the village to make a petition for disavowing his son and consider that Mr Dai does not be existed in his life") | 1 |
| | p23: ("ông/he", "sẽ ra/will go") | 0 |
| | p24: ("ông/he", "sẽ ra/will go", "làng, xã/the village") | 1 |
| | p25: ("ông/he", "sẽ ra/will go", "làm đơn từ con/for making a petition to disavow his son") | 1 |
| | p26: ("Ông/He", "quả_quyết/asserts", "thì ông sẽ ra làng , xã làm đơn từ con và coi như anh Đại không có trên cuộc_đời này/ that he will go to the village to make a petition for disavowing his son and consider that Mr Dai does not be existed in his life.") | 1 |
| | p27: ("ông/he", "sẽ ra/will go", "và coi như anh Đại không có trên cuộc_đời này/and consider that Mr. Dai does not be existed in his life.") | 0 |
| | p28: ("ông/he", "coi/do it") | 0 |
| | p29: ("anh Đại/Mr Dai, "không có/nothing") | 0 |
| | p30: ("anh Đại/Mr Dai", "không có/is not", "trên cuộc_đời này/in the life") | 1 |
| 6 | **Sentence 6:** Chị Huệ đưa tay chỉ cô bé ngồi bên cạnh: "Nếu không có bé Dung giúp_đỡ thì chắc_chắn tui đã cho Thương nghỉ học từ lâu rồi"./Ms Hue indicates the girl sitting nearby: "If there was not Dung's help, I could force Thuong leaving school." | |
| | p31: ("Chị Huệ/Ms. Hue", "đưa tay/indicates") | 1 |
| | p32: ("Chị Huệ/Ms. Hue", "đưa tay/indicates", "chỉ cô bé ngồi bên cạnh/the girl sitting nearby") | 1 |
| | p33: ("Chị Huệ/Ms. Hue", "chỉ cô bé ngồi bên cạnh/shows the girls sitting nearby") | 1 |
| | p34: ("Chị Huệ/Ms. Hue", "đưa tay/indicates", "Nếu không có bé Dung giúp_đỡ thì chắc_chắn tui đã cho Thương nghỉ học từ lâu rồi/ If there was not Dung's help, I could force Thuong leaving school.") | 0 |
| | p35: ("Chị Huệ/Ms. Hue", "không có/doesn't have") | 0 |
| | p36: ("Chị Huệ/Ms. Hue", "đưa tay/indicates", "bé Dung/Dung giúp_đỡ/help") | 0 |
| | p37: ("bé Dung/Dung", "giúp_đỡ/help") | 1 |
| | p38: ("tui/I", "đã cho/forced", "Thương/Thuong", "nghỉ học từ lâu rồi/leaving school for long time") | 1 |
| | p39: ("Chị Huệ/Ms. Hue", "đưa tay/indicates", "thì chắc_chắn tui đã cho Thương nghỉ học từ lâu rồi/ I could force Thuong leaving school.") | 0 |
| | p40: ("tui/I", "nghỉ học/leave school") | 0 |
| | p41: ("tui/I", "nghỉ học/leave school", "từ lâu rồi/for long time") | 0 |

Table 5 shows several sample outputs by the vnOIE system, in the increasing order of sentence structure complexity (or the number of verbs in a sentence from one to six). We observe that the complexity of sentence structure will affect the result of the vnOIE system based on DP analysis. A highly complex sentence structure will lead to an uninformative and incoherent DP analysis that will cause incorrect extractions in the system. Most of the extractions in cases #1 to #3 are correct relations because vnOIE can easily determine the relation between a subject, a predicate and their constituents based on clause structures via DP analysis. In complex cases such as cases #4 to #6, p13, p15, p16, p35, p40 and p41 are incorrect extractions due to weak connections between subjects and predicates. For example., in sentence 4, a relationship between the subject: "Má/Mama", verb: "xách/carry", and object: "giỏ/a bag" are shown in DP via *vmod*(phải/must, xách/carry) and *dob*(xách/carry, giỏ/bag) where "giỏ/bag" is considered to be averb auxiliary, which results in a wrong extraction in p16. In other cases, when DP shows poor relations between verbs and required objects or adverbs in a complex sentence for extracted propositions (*arg1*, *rel*, *arg2*), the system will fail to extract correct relations. As a consequence, the system extracted incorrect relations in p27, p28, p36 and p39 due to wrong relations between verbs and objects, and incorrect relations in p19, p23, and p29 due to uninformative relations. We hope that future improvements of Vietnamese DP with more detailed grammar analysis will help to improve the performance of our system.

## 5 CONCLUDING REMARKS

In this paper, we present vnOIE, a system for Vietnamese OIE. To the best of our knowledge, our system is the first contribution to the area of Vietnamese OIE, and takes advantages of the grammatical clause-based approach. In particular, we exploit Vietnamese DP using grammatical clauses, which strives to consider all possible relations in a sentence. In each clause, the corresponding clause type according to the grammatical function of its coherent constituent is determined as an extractable relation. As a result, our system can generate open relations and their arguments from Vietnamese text with highly scalable extraction while being domain independent. In our experiments, we evaluated the system using several factors such as grammatical structures of sentences and the number of verbs existing in a sentence. The results show that our system delivers promising results. Our system can further be applied to Vietnamese answering systems or integrated in higher levels of Vietnamese NLP tasks such text similarity or text summarization.